\pdfoutput=1
\documentclass[11pt]{article}

\usepackage[final]{coling}
\usepackage{times}
\usepackage{latexsym}
\usepackage{comment}
\usepackage{amsmath}
\usepackage{todonotes}
\usepackage{url}
\usepackage{cleveref}

\usepackage[T1]{fontenc}
\usepackage{tcolorbox}
\newtcolorbox{exampleblock}[1][]{
  colback=brown!20!white, colframe=brown!75!black, 
  fonttitle=\bfseries, title=LLM Prompt~\thetcbcounter, #1
}
\usepackage[utf8]{inputenc}

\usepackage{microtype}
\usepackage{inconsolata}
\usepackage{graphicx}

\title{Contextual Augmentation for Entity Linking using Large Language Models}

\author{Daniel Vollmers, Hamada M. Zahera, Diego Moussallem, Axel-Cyrille Ngonga Ngomo\\ 
  Data Science Group, Paderborn University, Germany \\
  \texttt{\{daniel.vollmers,hamada.zahera,diego.moussallem,axel.ngonga\}@uni-paderborn.de} 
  }

\begin{document}
\maketitle
\begin{abstract}
Entity Linking involves detecting and linking entity mentions in natural language texts to a knowledge graph. Traditional methods use a two-step process with separate models for entity recognition and disambiguation, which can be computationally intensive and less effective. We propose a fine-tuned model that jointly integrates entity recognition and disambiguation in a unified framework. Furthermore, our approach leverages large language models to enrich the context of entity mentions, yielding better performance in entity disambiguation. We evaluated our approach on benchmark datasets and compared with several baselines. The evaluation results show that our approach achieves state-of-the-art performance on out-of-domain datasets.
\end{abstract}

\section{Introduction}

%intr
Entity Linking (EL) in knowledge graphs (KGs) involves identifying and connecting entities within a text to their corresponding entries, enhancing the semantic understanding of the text~\cite{OLIVEIRA2021101624}. This process includes two main steps: (i) Named Entity Recognition (NER), which detects entity spans such as names, dates, and locations; and (ii) Entity Disambiguation (ED), which resolves ambiguities by accurately matching these entities to their corresponding entries in the knowledge graph. 

Traditional EL approaches, such as the two-stage architecture~\cite{el_survey}, divide the task into candidate generation and entity re-ranking phases. 
Another approach involves bi-encoder and cross-encoder models, as demonstrated by systems like BLINK~\cite{blink}. 
Bi-encoder models independently encode mentions and entities for efficient retrieval. In contrast, cross-encoder models jointly evaluate mention-entity pairs to enhance accuracy. Additionally, generative models~\cite{wang2023benchmarking} consider candidate generation as a text generation task, where models learn to generate unique entity names based on contextual information. However, 
While these traditional methods excel in identifying and linking common entities, they often struggle with handling long-tail entities, i.e., rare or those with multiple meanings, making it difficult to accurately disambiguate them in different contexts. 
For example, consider the entity \texttt{‘Jaguar’}. In a general context, \texttt{‘Jaguar’} could refer to the animal, the car brand, or even a sports team. However, in a domain-specific context, such as a biology research paper, \texttt{‘Jaguar’} would most likely refer to the animal. Traditional methods may not effectively handle such distinctions, leading to potential errors in Entity Linking. Another example, \texttt{"Angelina met her partner Brad and her father Jon in AK"}, where entities are identified by their first names, while news articles commonly use surnames.

In this paper, we propose a LLM-based augmentation strategy to enrich the context of entities mentions in short texts such as questions. Our approach expands entities mentions by prompting the LLM to extend them to their likely Wikipedia titles, thereby replacing ambiguous entity spans with more easily linkable ones. For example, the spans \texttt{“Angelina”}, \texttt{“Brad”} and \texttt{“Jon”} should be expanded to \texttt{“Angelina Jolie”}, \texttt{“Brad Pitt”} and \texttt{“Jon Voight”}. Additionally, our strategy  replaces abbreviations like \texttt{“AK”} commonly used for the state of Alaska, with the full name \textit{“Alaska”}. 
To link entities in out-of-domain datasets, we use an autoregressive model~\cite{genre}. which generates entity names token-by-token based on context. 
This approach allows the model to adapt to new or unseen entities by leveraging the surrounding text, thereby improving accuracy of identifying and linking entities that are absent from the training data. 
We conducted several experiments on different benchmarks to evaluate the performance of our approach against various baselines. 
Specifically, we experimented with two types of models: An end-to-end model, which directly links entities and a traditional two-step approach, which first identifies entity spans before applying the disambiguation step. 
The evaluation results demonstrate that our approach significantly outperforms different baselines by a large performance margin on the benchmarking datasets. We summarize the main contributions in this paper as follows: 
\begin{itemize}
       
    \item We propose an LLM-based approach for Entity Linking that leverages zero-shot prompting, which achieves state-of-the-art results on most out-of-domain evaluation datasets.
    
    \item We evaluated different LLM-based augmentation strategies for Entity Linking, comparing their effectiveness in both the two-step approach and the end-to-end approach.
    
    \item We evaluated the performance of a joint model for entity recognition and disambiguation compared to end-to-end models and NER models.
    
    \item We make our source code and fine-tuned models publicly available at the GitHub repository.\footnote{\url{https://github.com/dice-group/AugmentedEL}}
    
\end{itemize}

\section{Related Work}
Entity Linking is typically involves two phases: \textit{span detection} and \textit{entity disambiguation}~\cite{el_survey}. 
During span detection phase, most existing approaches employ standard named entity recognition to identify relevant spans in text. In the disambiguation phase, candidate entities are generated from a knowledge graph and linked to the most appropriate match using pre-built search indices, where each entry corresponds to a KG entity. 

\paragraph{Disambiguation}
Approaches like MAG \cite{MAG} and DoSeR \cite{doser} rely on pre-built indices for effective entity disambiguation. 
For example, MAG utilizes five distinct indices—\textit{surface forms, personal names, rare references, acronyms}, and \textit{contextual information}—to query entities. 
Following the index query, MAG applies an additional step to refine the candidate set for final disambiguation. In contrast, DoSeR integrates text-based retrieval with surface forms, leverages a Word2Vec embedding model, and uses a priori probabilities derived from occurrence frequency for candidate generation, employing a similar candidate expansion strategy as MAG. 
Other approaches such as \citet{Mulang__2020} introduce context information by extracting triples from knowledge graphs and verbalizing them  to the input sequence. Meanwhile, \citet{tplink} incorporate type information into the disambiguation process to improve accuracy. 
Moreover, 
BLINK \cite{blink} adopts an embedding-centric approach for candidate generation, employing a Bi-Encoder model to generate representations of both candidates and mentions. Entity search within the index is executed through KNN-Search based on context embedding vectors. Other methods, similar to BLINK \cite{tuanmanh}, combine text-based retrieval with deep neural embedding models for entity disambiguation. 

Alternatively, \citet{parravicini} propose an embedding-based approach where node embeddings, derived using the \textit{word2vec} algorithm, assesses vertex similarity. This method involves evaluating candidates via tuples, where each tuple corresponds to candidate entities linked to mentions in a document. A global similarity score, calculated from these node embeddings, determines the score for each tuple.
In contrast, some approaches utilize re-ranker models to calculate embeddings, considering both the mention's context and candidate entities. These models employ a feedforward layer to re-rank candidates~\cite{blink,tuanmanh}. Lastly,\citet{xin2024llmaellargelanguagemodels} introduce the first use of LLMs for context augmentation in entity disambiguation, focusing solely on enhancing the context of entity mentions without addressing NER. This method, however, is computationally intensive due to the need to augment each mention individually.

 \begin{figure*}[t!]
    \centering
    \includegraphics[trim={0.6cm 0.6cm 0.6cm 0.6cm},clip,width=\textwidth]{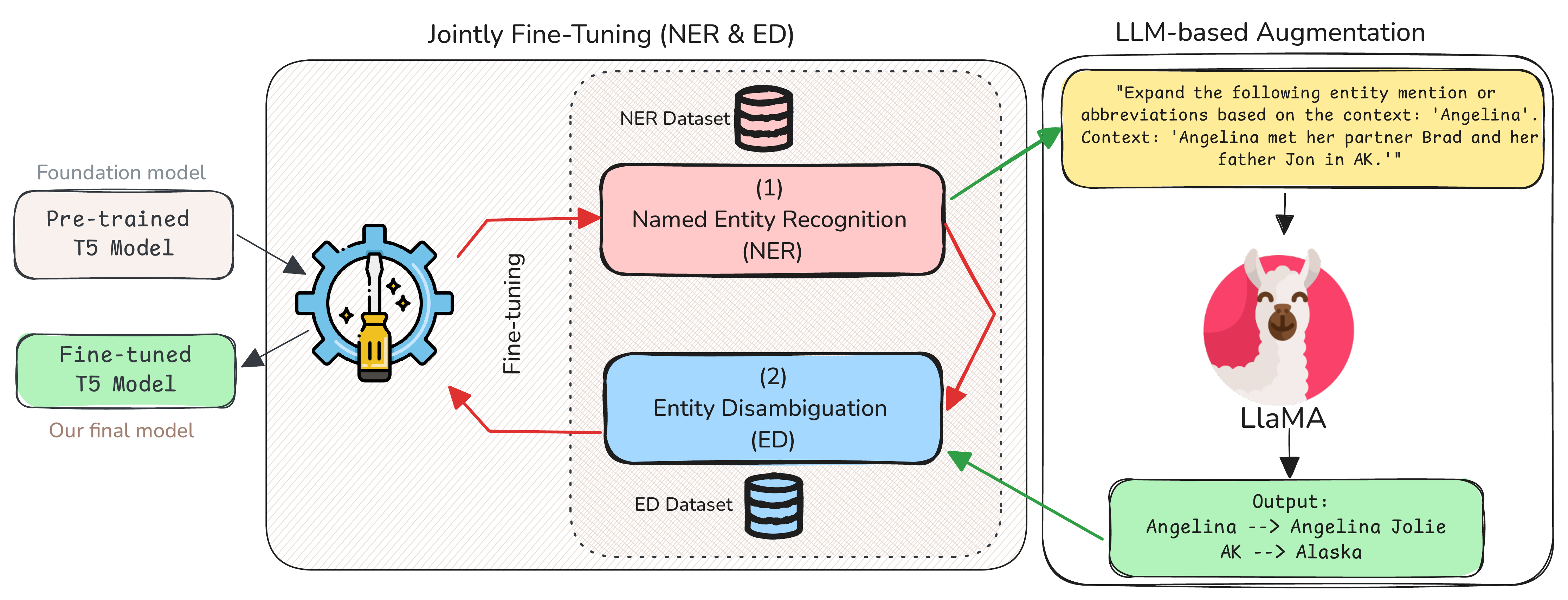}
    \caption{The architecture of our approach, including Jointly Fine-tuning and LLM-based Augmentation}
    \label{fig:model}
\end{figure*}
\paragraph{End-to-end Entity Linking}
Unlike, the traditional two-step Entity Linking process, recent approaches omit the NER step and directly extracting or annotating entity candidates from the input sequence. These approaches often employ fine-tuned models for autoregressive annotation~\cite{DBLP:journals/corr/abs-2110-02369}. 
Early work by~\citet{kolitsas-etal-2018-end} integrated mention and entity embeddings based on Word2Vec and sequence-to-sequence models like LSTM \cite{lstm} to incorporate contextual information. 
Furthermore,~\citet{rel} introduced an approach that predicts coherence scores to align entity annotation. Modern Entity Linking methods mainly involve fine-tuning Large LLMs for this task. For instance, GENRE framework \cite{genre} leverages a fine-tuned BART model for autoregressive annotation of input sequences, utilizing an offline prefix trie from Wikipedia titles to constrain the decoding process and thus reduce the search space. In contrast, EntQA approach~\cite{DBLP:journals/corr/abs-2110-02369} uses a vector-based search index similar to BLINK to identify entities within the input sequence. During the disambiguation phase, it computes candidate spans for each entity and selects the highest-scored candidate for linking. 
Our work differs from these methods by addressing the entire Entity Linking pipeline, including NER, and introduces an LLM-based augmentation to improve entity disambiguation, particularly on out-of-domain datasets.

\section{Approach}
This section outlines our approach for Entity Linking using a fine-tuned T5 model and contextual augmentation using LLMs. 
First, we provide the task definition of Entity Linking, then a description our approach's with the main components, including: \textit{Named Entity Recognition (NER)}, \textit{LLM-based Augmentation}, \textit{Entity Disambiguation (ED)} and the \textit{Joint Fine-tuning of (NER\&ED)}. Afterwards, we describe our strategy to mitigate LLMs hallucination and the ablations of our approach to assess the performance of LLM-based augmentation.

\subsection{Task Definition}
Entity Linking task involves two main steps: (I) Named Entity Recognition, and (II) Entity Disambiguation.
Named Entity Recognition identifies and extracts entity mentions (e.g., people, organizations, locations) from text. Given a context $C= (t_1, t_2,\cdots, t_k)$, the goal is to identify a subset of tokens $M={m_1, m_2, \cdots, m_i}$ representing entity mentions. The NER function maps $C$ to a list of entity mention spans $M$ ~\cite{el_survey}.
\begin{equation}
\text{NER}:C \rightarrow M^n
\end{equation}

After identifying entities, Entity Disambiguation resolves any potential ambiguities by linking each mention $m_i$ to the correct entity $e_i$ in a knowledge graph, using similarity measures and contextual information~\cite{el_survey}.
\begin{equation}
\text{ED}:\left[(m_1, \cdots, m_n),C\right] \rightarrow (e_1, \cdots, e_n)
\end{equation}

\subsection{Architecture}
We employ the T5 model as the \textit{foundation architecture} in our approach and fine-tune it on NER and ED tasks. This joint fine-tuning allows our model to leverage shared knowledge across both tasks, improving its overall performance. The following sections summarize the main components of our approach.

\subsubsection{Named Entity Recognition}
\label{sec:ner}
We use the T5 model with a transformer architecture to captures contextual information from both preceding and succeeding text. Furthermore, the T5 model regards NER task as a text-to-text problem, where both input and output are sequences of text. 
Initially, the input text is tokenized and processed by the model’s encoder, which generates contextual embeddings by considering the surrounding words. These embeddings are then used by the decoder to produce an output sequence with entity tags. 

In our approach, we fine-tune the T5 model on annotated datasets with entities, allowing it to learn accurate tagging based on context. For instance, the sentence \texttt{"Angelina met her partner Brad and her father Jon in AK"} is transformed into \texttt{[\text{BEGIN\_ENT}] Angelina [\text{END\_ENT}] met her partner [\text{BEGIN\_ENT}] Brad [\text{END\_ENT}] and her father [\text{BEGIN\_ENT}] Jon [\text{END\_ENT}] in [\text{BEGIN\_ENT}] AK [\text{END\_ENT}]}. In this output, \texttt{"Angelina"}, \texttt{"Brad"}, \texttt{"Jon"}, and \texttt{"AK"} are recognized as entities and marked with an annotation tag.

\subsubsection{LLM-based Augmentation}
\label{sec:augmentation}
To further augment the NER process, we employ the LlaMA3 model to perform additional entity recognition and expand on the entities detected from the previous step. 
The core idea is to replace ambiguous or incomplete entity mentions with more precise and recognizable forms, such as full names or specific titles. 
This is achieved by prompting the LlaMA3 model to generate these extended forms based on the given context. Our prompt includes a structured input with the entity mention and its surrounding context. For example, consider the entity mention \texttt{"Angelina"} in the sentence: \textit{"Angelina met her partner Brad and her father Jon in AK."}. We prompt the LlaMA3 model to expand entity mentions as follows:
\begin{exampleblock}[label={ex:ample}]
\texttt{\textbf{Expand} the following entity mention 'Angelina' and abbreviations 'AK' based on the context: . 
\\
\textbf{Context}: 'Angelina met her partner Brad and her father Jon in AK.'}
\end{exampleblock}

In response, the LlaMA3 model expands the entities (e.g., \texttt{"Angelina"}$\longrightarrow$\texttt{"Angelina Jolie"} and for the abbreviation \texttt{"AK"}$\longrightarrow$\texttt{"Alaska"}. This approach allows us to replace ambiguous mentions with their more specific counterparts, thereby improving the ability of the model to link entities in the follow-up entity disambiguation step. 

\subsubsection{Entity Disambiguation}
\label{sec:disamb}
Entity Disambiguation is crucial for resolving ambiguities when multiple entities share the same name. In our approach, the T5 model addresses this by encoding the context around an entity mention and creating a representation that captures both semantic (e.g., word meaning) and syntactic information (e.g., sentence structure). 
determine the correct entity based on the surrounding text. Then, it decodes this representation to predict the correct entity from a set of candidates, using attention mechanisms to focus on relevant parts of the input text. Formally, given a context $C$, the model generates an output sequence $T$ with entity mentions $m \in M^n$ and their corresponding URIs $e \in E^n$, represented by their titles in a target knowledge graph. The output structure is defined as: 

\begin{equation}
[\text{BEGIN\_ENT}]~m~[\text{END\_ENT}] [\text{title}(e)]
\end{equation}
where $m \in M^n$ denotes an entity mention and $e \in E^n$ represents the associate URI. 

Unlike traditional methods, our approach is unique as it leverages one single T5 model to perform both NER and ED sequentially. First, the T5 model generates an intermediate sequence $I$ for NER:
\begin{equation}
\text{I} = [\text{BEGIN\_ENT}]~m~[\text{END\_ENT}] | m \in \text{M}^n
\end{equation}
Afterward, it predicts the target output based on the output of the NER step.
At inference time, the target output is expanded by our augmentation strategy as presented in section \ref{sec:augmentation}.

\subsubsection{Jointly Fine-tuning (NER\&ED)}
To fine-tune the T5 model for both NER and Entity Disambiguation, we create two different training samples: one for the NER task and another for the Disambiguation task. For the NER task, the input is a text sequence without annotations, appended with the suffix \textit{target\_ner}. The target sequence is the same text with annotated entity mentions, as detailed in section \ref{sec:ner}. For the Disambiguation task, the input is a text sequence with annotated entities, appended with the suffix \textit{target\_el}, and the target sequence includes the corresponding Wikipedia entity labels, as described in section \ref{sec:disamb}.

We combine the NER and Disambiguation samples into a single dataset to fine-tune the model jointly for both tasks. Additionally, we integrate the NER and Entity Disambiguation tasks within a unified framework (see~\Cref{fig:model}). This integration enhances the robustness and accuracy of our approach in identifying and disambiguating entities, leading to improved performance in Entity Linking.

\subsection{Mitigating LLMs Hallucination}
During our implementation, we found that LLM hallucination is a critical problem, especially in the augmentation and disambiguation phase. In the disambiguation step, the LLM model occasionally predicts labels, that do not exist in Wikipedia. 
To avoid this, we generated a dictionary that maps all Wikipedia titles to their URIs. By applying this dictionary, we can omit all annotations, where no exact match in the dictionary exists. 
In the augmentation step, hallucination occurs when the LLM model provides augmentations for spans that do not exist in the sequence or are not annotated by the NER step. 
To avoid misleading expansions, we only consider those spans for expansion that precisely match one of the annotated spans from the NER step. Unlike, the augmentation and disambiguation steps, we did not encounter problems with hallucination in the NER step.

\section{Ablations}
\label{sec:ablations}
We conducted various ablations of our model to evaluate the impact of augmentation in different setups of EL models, as follows:

\paragraph{End-to-end foundational model}
Recent studies focuses on developing Entity Linking models that omits mention detection step and directly predicts the entity set~\cite{genre,DBLP:journals/corr/abs-2110-02369}. In the end-to-end (E2E) approach, an expanded set of entities is used to directly compute $E^n$ for the context $C$.

To setup the end-to-end approach, we trained our T5 model to directly predict the target sequence $T$ from the input context $C$. 
This approach aligns with the same experiments setup by~\citeauthor{genre} on end-to-end Entity Linking. 
However, our implementation employs the T5-base model instead of the BART model. 
We selected the T5-based model due to its superior performance without requiring a prefix trie. 
Additionally, we used an augmentation strategy with two inference steps with the model: first, the model identifies entity mentions to be expanded, then, performs  final entity disambiguation. 
After the first step, we extract entities mentions from the output sequence, ignoring the predicted titles. Subsequently, we use the same LLM prompt, as in the augmentation~\Cref{ex:ample}, to find possible expansions. To integrate these expansions into the sequence, we replace the mentions with their expanded forms, omitting the tags for annotating entity mentions, similar to the foundational model.

\paragraph{NER augmentation} 
To improve the performance of our Entity Linking system, we introduce our LLM prompt to find additional entity spans in the input texts. We use the following instruction to guide the LLM in finding accurate entity spans from the input text:
\begin{exampleblock}[label={example_ner}]
\texttt{Please \textbf{generate one list with all entities} from the following text in JSON format, excluding numbers. Do not format the json output: \\
\textbf{Context}: 'Angelina met her partner Brad and her father Jon in AK.'}
\end{exampleblock}

We then apply a regular expression, similar to the first expansion strategy, to extract the new entity spans. 
Since we cannot rely on specified indices in the LLM output, we only consider spans that exactly match the original input sequence. 
To avoid overlapping annotations, we order the newly extracted spans by length in descending order and add expansions only where no surrounding annotation is present.

\paragraph{Alternative NER approaches}
\label{found_models}
we conducted experiments using the state-of-the-art framework Flair~\cite{akbik2019flair} for the mention detection step.
Additionally, we experimented with a hybrid approach that employs an end-to-end foundational model for mention detection and then introduces those entities into the disambiguation step.

\section{Experiments}
We conducted our experiments to answer the following research questions:
\begin{itemize}
    \item[\textbf{RQ}$_1$] How well does our approach compared to state-of-the-art baselines?
    \item[\textbf{RQ}$_3$] How does the LLM-based augmentation impact foundational models?
    \item[\textbf{RQ}$_3$] Which augmentation strategy works best?
\end{itemize}

\subsection{Experimental Setup}
We fine-tuned the T5 model, using it as a foundational model, on the KILT dataset. The fine-tuning was conducted on 4 NVIDIA-A100 GPUs, each with 40GB of memory, for one week. Following this, we further trained the model on the AIDA-train dataset for up to 125 epochs with an early stopping strategy. We used the AIDA-test-A split as the development set.
We chose the base version of the T5 model due to its size, which is comparable to the models used in baseline approaches. The T5 implementation was obtained from Hugging Face.\footnote{\url{https://huggingface.co/docs/transformers/model_doc/t5}}
Our training setup mirrors that of the baseline approaches; no additional datasets were used for further training. The end-to-end foundational model, as detailed in Section~\ref{sec:ablations}, was trained under the same conditions.
For inference, we deployed the foundational model and the LLM for the augmentation strategy on a separate machine equipped with two NVIDIA H100 GPUs. This setup facilitated the use of large models like the LLaMA3 with 70 billion parameters.

\subsection{Evaluation}
We conducted our experiments using the GERBIL framework~\cite{gerbil}, which benchmarks Entity Linking on various datasets. We configured an A2KB experiment and integrated our approach as a web service. We report the InKB micro F1 scores, which is a standard metric in the literature. This metric considers only entities with a corresponding link in the target knowledge graph, thereby excluding \textit{out-of-wiki} entities from the evaluation~\cite{gerbil}. For our baseline experiments, we focused only on the approaches that were evaluated in an end-to-end setup, including NER. This inclusion is critical as the NER output significantly impacts the performance of entity disambiguation.

\subsection{Datasets}
We used different datasets for the end-to-end Entity Linking task in our experiments~\cite{genre,DBLP:journals/corr/abs-2110-02369}. These datasets are: AIDA-test-B~\cite{hoffart-etal-2011-robust}(AIDA), Derczynski~\cite{Derczynski_2015}(DER), KORE 50~\cite{k50}(K50), MSNBC, NS3-Reuters-128, NS3-Reuters-500 ~\cite{roder-etal-2014-n3}(R-128,R-500), and the OKE challenge datasets OKE-2015 and OKE-2016~\cite{Nuzzolese2015OpenKE}. \Cref{tab:datasetstatistics} provides detailed statistics of these datasets, including number of entities that have a corresponding entry in the knowledge graph (\#InKG entities) and number of documents (\#Docs). 
The AIDA-test-B~\cite{hoffart-etal-2011-robust} dataset contains the largest number of entities with corresponding entries in the knowledge graph. Given that our models were trained on the AIDA-train dataset, AIDA-test-B serves as an in-domain dataset. All other datasets are considered out-of-domain. For the knowledge graph, we used the 2019-Wikidata-dump from the KILT dataset,~\footnote{\url{https://github.com/facebookresearch/KILT}} which is commonly used by state-of-the-art Entity Linking systems.

\begin{table}[t!]
    \setlength{\tabcolsep}{6pt}
	\centering	
 	\caption{Dataset Statistics}
	\begin{tabular}{lrrrr} 
		\hline
		\textbf{Dataset}& \textbf{\#InKB entities} & \textbf{\#Docs}    \\ 
		\hline
		AIDA-test-B & 4,485&230  \\
        Der&201&183\\
        KORE50 & 139&48  \\
        MSNBC & 737& 20 \\
        N3-Reuters-128 & 626 &115 \\
        N3-RSS-500 & 515 &425 \\
        oke-2015 & 481 & 100\\
        oke-2016 & 221 & 55\\
		\hline
	\end{tabular}

  \label{tab:datasetstatistics}
\end{table}

\subsection{Experimental Results}
\subsubsection{Comparison to Baseline Approaches \textbf{(RQ)}$_1$}
\label{sec:baseline_comp}

\begin{table*}[t!]
\centering
\caption{Performance comparison against baseline approaches using InKB micro F1 (\textbf{RQ}$_1$). T5$_{(\text{ED})}$ is fine-tuned for Entity Disambiguation, while T5$_{(\text{NER+ED})}$ is jointly fine-tuned for both NER and Disambiguation. The best result is highlighted in bold.}
\resizebox{\textwidth}{!}{
\begin{tabular}{@{}lcccccccc|c@{}}
\hline
\textbf{Approach} & \multicolumn{1}{l}{\textbf{AIDA}} &\multicolumn{1}{l}{\textbf{MSNBC}}&\multicolumn{1}{l}{\textbf{Der}}&\multicolumn{1}{l}{\textbf{K50}}&\multicolumn{1}{l}{\textbf{R-128}}&\multicolumn{1}{l}{\textbf{R-500}}&\multicolumn{1}{l}{\textbf{OKE 2015}}&\multicolumn{1}{l}{\textbf{OKE 2016}} &\multicolumn{1}{l}{\textbf{Avg}} \\ \hline
\citeauthor{hoffart-etal-2011-robust}   & 72.8& 65.1& 32.6& 55.4& 46.4& 42.4& 63.1& 0.00 &47.2 \\
\citeauthor{DBLP:conf/esws/SteinmetzS13} &  42.3 &30.9 &26.5& 46.8& 18.1 &20.5& 46.2 &46.4 &34.7\\
\citeauthor{moro-etal-2014-entity} & 48.5 &39.7& 29.8 &55.9 &23.0& 29.1& 41.9 &37.7 &38.2\\
\citeauthor{kolitsas-etal-2018-end}  &  82.4& 72.4& 34.1& 35.2& 50.3 &38.2 &61.9& 52.7 &53.4\\
\citeauthor{rel}  & 80.5& 72.4& 41.1 &50.7& 49.9 &35.0 &63.1 &58.3 &56.4\\
\citeauthor{genre}   &  83.7 &\textbf{73.7}& 54.1& 60.7& 46.7 &40.3& 56.1& 50.0 &58.2\\
\citeauthor{DBLP:journals/corr/abs-2110-02369}  &  \textbf{85.8} &72.1& 52.9& 64.5& 54.1& 41.9& 61.1 &51.3 &60.5\\
\hline
Flair \& T5$_{(\text{ED})}$ &71.4 &61.3&51.6 &\textbf{72.7} &\textbf{54.5}& 56.3& \textbf{66.6} &\textbf{61.5} &\textbf{62.0}\\
T5$_{(\text{NER+ED})}$  & 71.6& 69.3& \textbf{55.7}& 70.6& 51.7& \textbf{56.6}& 59.4& 58.5 &61.7\\
E2E T5 & 69.0  &64.2&53.7 &64.3 &51.9& 57.3& 61.6 &58.4 &60.1\\
\hline
\end{tabular}
}
\label{tab:baselines}
\end{table*}

In this section, we compared the performance of our model against state-of-the-art baselines across various datasets. We obtained the scores of baselines from~\citeauthor{DBLP:journals/corr/abs-2110-02369}. Our evaluation focused on three setups that demonstrated the most stable results: the foundational T5 model with entity span expansion (T5$_{(NER+ED)}$), a combination of the T5 model with the Flair framework for NER (Flair \& T5$_{(ED)}$), and an end-to-end T5 ablation model (\Cref{sec:ablations}) with span expansion (E2E T5).

\Cref{tab:baselines} reports the evaluation results on across all datasets. 
Unlike traditional Entity Linking systems such as~\cite{hoffart-etal-2011-robust} and~\cite{rel}, which uses separate components for span detection and entity disambiguation, our models do not generate candidate entity sets. Instead, we use the fine-tuned T5 models with LLM-augmentation for entity expansions for contextual information.
Our model achieves the best performance on all datasets except AIDA and MSNBC. 
On the MSNBC dataset, our model's performance is comparable to the state-of-the-art. However, we observed a performance drop on the in-domain AIDA-test-B dataset compared to the baselines. Interestingly, the expansion strategy is less effective on the AIDA and MSNBC datasets, as described in~\Cref{sec:eval_foundational}. This might be due to our foundational models have been trained the AIDA data, so the expansion did not provide substantial new contextual information for the input sequences. 
The MSNBC dataset, which includes 20 news articles, shares similarities with the in-domain AIDA-test-B dataset. 
The other datasets often contain much shorter texts with fewer entities and are exclusively news-based. 

Our findings indicate that the more a dataset differs from our training data, the better our model performs relative to the baselines. Notably, on the KORE 50 dataset, which contains highly ambiguous entities, our model with a joint mention detection and disambiguation improves the F1 score by over $6$ points  and $8$ points when combined with the third-party flair NER framework.

\subsubsection{Evaluation of foundational models \textbf{(RQ)}$_2$}
\label{sec:eval_foundational}
To address this research question, we evaluated the performance of various foundational models, specially T5, and investigated how our LLM-based augmentation strategy on affected their performances. We employed the LlaMA3 model for entity expansion (e.g., \texttt{"Angelina" to "Angelina Jolie"}) to facilitate  entity disambiguation, as described in~\Cref{sec:expansion_strategies}). 

\Cref{tab:found_models} reports the evaluation results for this experiment. 
We employed the same models as in the previous section, along with the mixed foundational model (Mixed model) presented in~\ref{found_models}. 
By analyzing the results of both the foundational models and those enhanced with our expansion strategy, we observed significant improvements on most out-of-domain datasets. 
The augmented version of the traditional setup, which separates mention detection and evaluation-- outperforms the augmented end-to-end model on five out of eight datasets. 
For the other three datasets, the performance difference between the two setups was negligible. However, without augmentation, the end-to-end model generally performs works better than the traditional setup.
This finding suggests that our LLM-based entity expansion is particularly effective with a traditional setup for disambiguating and liking entities to knowledge graphs more accurately. 

\begin{table*}[t!]
\centering
\caption{Comparison of different foundational models \textbf{(RQ)}$_2$. The first 4 rows present of the approach, when no augmentation strategy is applied. The other rows present results when the entity expansion strategy (\Cref{sec:augmentation}) is applied. The best result is highlighted in bold.}
\resizebox{\textwidth}{!}{
\begin{tabular}{@{}lcccccccc|c@{}}
\hline
\textbf{Model}          & \multicolumn{1}{l}{\textbf{AIDA}} &\multicolumn{1}{l}{\textbf{MSNBC}}&\multicolumn{1}{l}{\textbf{Der}}&\multicolumn{1}{l}{\textbf{K50}}&\multicolumn{1}{l}{\textbf{R-128}}&\multicolumn{1}{l}{\textbf{R-500}}&\multicolumn{1}{l}{\textbf{OKE 2015}}&\multicolumn{1}{l}{\textbf{OKE 2016}} & \multicolumn{1}{l}{\textbf{Avg}} \\ \hline
Flair \& T5$_{(\text{ED})}$  &\textbf{73.5} &67.0 &48.8 &50.0 &41.0&55.1 & 58.9& 55.0& 56.2\\
T5$_{(\text{NER+ED})}$   &67.4 & 61.5&53.9 &50.0 &47.6&55.2 &55.9 &56.0& 56 \\
E2E T5 &68.0   &63.0 &50.4& 49.6&49.1  &56.8&56.1  &56.5 &56.1\\
Mixed model &66.8  &61.9& 53.5 & 48.5&49.2&55.3 &56.8  &50.0 &55.2 \\
\hline
Flair \& T5$_{(\text{ED})}$ aug. &71.4 &61.3&51.6 &\textbf{72.7} &\textbf{54.5}& 56.3& \textbf{66.6} &\textbf{61.5} & \textbf{62.0}\\
T5$_{(\text{NER+ED})}$ aug.  & 71.6& \textbf{69.3}& \textbf{55.7}& 70.6& 51.7& 56.6& 59.4& 58.5 & 61.7\\
E2E T5 aug.&69.0  &64.2&53.7 &64.3 &51.9& 57.3& 61.6 &58.4 &60.1 \\
Mixed model aug.  &  69.0& 63.1&55.0 &69.6 &52.3  &\textbf{58.1} &60.6&53.5 &60.2 \\
%GrailQA s-expressions& 0.31&0.35& 0.33 &0.50 \\
\hline
\end{tabular}
}
%\vspace{0.5em}

\label{tab:found_models}
\end{table*}

\subsubsection{Evaluation of augmentation strategies \textbf{(RQ)}$_3$}
\begin{table*}[ht!]
\centering
\caption{Evaluation of different augmentation strategies (\textbf{RQ}$_3$). Models with NER-exp indicate the use of both NER and mention expansion, while other models use only mention expansion. The best result is highlighted in bold.}
\resizebox{\textwidth}{!}{
\begin{tabular}{@{}lcccccccc|c@{}}
\hline
\textbf{Model}          & \multicolumn{1}{l}{\textbf{AIDA}} &\multicolumn{1}{l}{\textbf{MSNBC}}&\multicolumn{1}{l}{\textbf{Der}}&\multicolumn{1}{l}{\textbf{K50}}&\multicolumn{1}{l}{\textbf{R-128}}&\multicolumn{1}{l}{\textbf{R-500}}&\multicolumn{1}{l}{\textbf{OKE 2015}}&\multicolumn{1}{l}{\textbf{OKE 2016}} &\multicolumn{1}{l}{\textbf{Avg}} \\ \hline
T5$_{(\text{NER+ED})}$  & \textbf{71.6}& 69.3& \textbf{55.7}& 70.6& 51.7& \textbf{56.6}& 59.4& 58.5 &61.7\\
T5$_{(\text{NER+ED})}$, NER-exp  & 68.0 & \textbf{70.3}&37.4 &69.3 & 53.0 &45.4 &44.6&\textbf{57.4} &55.7\\
Flair \& T5$_{(\text{ED})}$   &71.4 &61.3&51.6 &72.7 &\textbf{54.5}&56.3& \textbf{66.6} &61.5 &\textbf{62.0}\\
Flair \& T5$_{(\text{ED})}$, NER-exp.& 67.4 & 57.0&37.1 &76.1& 51.7 &46.5 &62.8&61.2 &57.5\\
LLM-only&63.7&62.6&36.4&\textbf{75.1}&49.3&47.3&61.7&59.6&57.0\\
%GrailQA s-expressions& 0.31&0.35& 0.33 &0.50 \\
\hline
\end{tabular}
}
%\vspace{0.5em}
\label{tab:expansion strategies}
\end{table*}

\label{sec:expansion_strategies}
\Cref{tab:expansion strategies} shows the results for the two augmentation strategies --mention expansion and NER expansion-- for both the foundational model (T5$_{(\text{NER+ED})}$) and the combination of Flair and the foundational model (Flair \& T5$_{(\text{ED})})$. 
As previously described, the mention expansion consistently improves the performance of the Entity Linking system. 
Conversely, the NER expansion strategy shows variable efficacy on some of the datasets. It improves performance on some datasets but leads to worse results on others compared to using only mention expansion. This variation in performance may originate from the differences in annotation strategies among the datasets.
Particularly, in the Der dataset, there is a significant variance in performance, where the NER expansion strategy annotates more entities the benchmarks datasets, negatively impacting results relative to the foundational model and other strategies. 

To further explore this issue, we conducted an additional experiment using only the LLM expansion strategy for entity extraction, labeled as "LLM-only" in~\Cref{tab:expansion strategies}). 
The results indicate that the LLM-only strategy performs well with short texts in the KORE 50 dataset but underperforms on other datasets when not combined with with the T5 model. 
We retain the T5 model for predicting final identifiers, since using mention expansion alone led to hallucinations, making many predicted identifiers untraceable in the title dictionary.

\section{Conclusion and future work}
In this paper, we present our approach for Entity Linking using a jointly fined-tuned model and contextual augmentation with LLMs. 
In particular, our approach employs a fine-tuned T5 model that integrates NER and disambiguation tasks into a unified framework, reducing resource demands compared to separate models for each task. 
Although this setup may slightly drop performance, our experiments showed that this performance loss is only marginal, mainly due to unseen entities. 
Furthermore, our approach leverages the LlaMA-3-70B model to expend entities mentions with contextual augmentation. The evaluation results demonstrate that LLM-based augmentation significantly improves the performance on out-of-domain datasets, achieving state-of-the-art results compare to traditional two-step methods (i.e., entity recognition and disambiguation). 

Future work will focus on analyzing the performance of LLM-based disambiguation strategies in predicting rare entities, which require creating a new benchmark datasets. In the appendix, we present an additional experiment that shows larger LLMs performs better than smaller ones, as they return more consistent output with fewer variations. 

\section{Limitations}
Similar to the approaches~\cite{genre}, we use Wikipedia as the main knowledge graph, where unique titles facilitate entity identification. However, this method is not always possible with other knowledge graphs like Wikidata~\cite{wikidata}. 
Existing benchmarks and training datasets are also based on Wikipedia, making text-based features sufficient for efficient Entity Linking. 
However, in other knowledge graphs, the graph-based structure is crucial for disambiguation. For instance, the German state of Berlin and the city of Berlin share the same label but are distinct entities, making interconnected entities crucial for disambiguation. The current benchmarking datasets used for evaluation are outdated and do not address newer challenges like predicting rare entities. Additionally, it is unclear to what extent these datasets have been included in the training data of current LLMs. Therefore, future research should focus on developing new datasets that also address the limitations of modern LLMs.

\section*{Acknowledgement}
This work has been supported by the Ministry of Culture and Science of North Rhine-Westphalia (MKW NRW) within the project SAIL under the grant no NW21-059D, by the German Federal Ministry of Education and Research (BMBF) within the project KIAM under the grant no 02L19C115, by the German Federal Ministry of Education and Research (BMBF) within the project COLIDE under the grant no 01IS21005D and the European Union’s Horizon Europe research and innovation programme under grant agreement No 101070305.
\bibliography{custom}

\begin{thebibliography}{26}
\providecommand{\natexlab}[1]{#1}

\bibitem[{Akbik et~al.(2019)Akbik, Bergmann, Blythe, Rasul, Schweter, and
  Vollgraf}]{akbik2019flair}
Alan Akbik, Tanja Bergmann, Duncan Blythe, Kashif Rasul, Stefan Schweter, and
  Roland Vollgraf. 2019.
\newblock {FLAIR}: An easy-to-use framework for state-of-the-art {NLP}.
\newblock In \emph{{NAACL} 2019, 2019 Annual Conference of the North American
  Chapter of the Association for Computational Linguistics (Demonstrations)},
  pages 54--59.

\bibitem[{{De Cao} et~al.(2021){De Cao}, Izacard, Riedel, and Petroni}]{genre}
Nicola {De Cao}, Gautier Izacard, Sebastian Riedel, and Fabio Petroni. 2021.
\newblock \href {https://openreview.net/forum?id=5k8F6UU39V} {Autoregressive
  entity retrieval}.
\newblock In \emph{9th International Conference on Learning Representations,
  {ICLR} 2021, Virtual Event, Austria, May 3-7, 2021}. OpenReview.net.

\bibitem[{Derczynski et~al.(2015)Derczynski, Maynard, Rizzo, van Erp, Gorrell,
  Troncy, Petrak, and Bontcheva}]{Derczynski_2015}
Leon Derczynski, Diana Maynard, Giuseppe Rizzo, Marieke van Erp, Genevieve
  Gorrell, Raphaël Troncy, Johann Petrak, and Kalina Bontcheva. 2015.
\newblock \href {https://doi.org/10.1016/j.ipm.2014.10.006} {Analysis of named
  entity recognition and linking for tweets}.
\newblock \emph{Information Processing \&amp; Management}, 51(2):32–49.

\bibitem[{Hochreiter and Schmidhuber(1997)}]{lstm}
Sepp Hochreiter and J\"{u}rgen Schmidhuber. 1997.
\newblock \href {https://doi.org/10.1162/neco.1997.9.8.1735} {Long short-term
  memory}.
\newblock \emph{Neural Comput.}, 9(8):1735–1780.

\bibitem[{Hoffart et~al.(2012)Hoffart, Seufert, Nguyen, Theobald, and
  Weikum}]{k50}
Johannes Hoffart, Stephan Seufert, Dat~Ba Nguyen, Martin Theobald, and Gerhard
  Weikum. 2012.
\newblock \href {https://doi.org/10.1145/2396761.2396832} {{KORE}: keyphrase
  overlap relatedness for entity disambiguation}.
\newblock In \emph{Proceedings of the 21st ACM International Conference on
  Information and Knowledge Management}, CIKM '12, page 545–554, New York,
  NY, USA. Association for Computing Machinery.

\bibitem[{Hoffart et~al.(2011)Hoffart, Yosef, Bordino, F{\"u}rstenau, Pinkal,
  Spaniol, Taneva, Thater, and Weikum}]{hoffart-etal-2011-robust}
Johannes Hoffart, Mohamed~Amir Yosef, Ilaria Bordino, Hagen F{\"u}rstenau,
  Manfred Pinkal, Marc Spaniol, Bilyana Taneva, Stefan Thater, and Gerhard
  Weikum. 2011.
\newblock \href {https://aclanthology.org/D11-1072} {Robust disambiguation of
  named entities in text}.
\newblock In \emph{Proceedings of the 2011 Conference on Empirical Methods in
  Natural Language Processing}, pages 782--792, Edinburgh, Scotland, UK.
  Association for Computational Linguistics.

\bibitem[{Kolitsas et~al.(2018)Kolitsas, Ganea, and
  Hofmann}]{kolitsas-etal-2018-end}
Nikolaos Kolitsas, Octavian-Eugen Ganea, and Thomas Hofmann. 2018.
\newblock \href {https://doi.org/10.18653/v1/K18-1050} {End-to-end neural
  entity linking}.
\newblock In \emph{Proceedings of the 22nd Conference on Computational Natural
  Language Learning}, pages 519--529, Brussels, Belgium. Association for
  Computational Linguistics.

\bibitem[{Lai et~al.(2022)Lai, Ji, and Zhai}]{tuanmanh}
Tuan~Manh Lai, Heng Ji, and ChengXiang Zhai. 2022.
\newblock \href {https://doi.org/10.48550/ARXIV.2202.13404} {Improving
  candidate retrieval with entity profile generation for wikidata entity
  linking}.
\newblock \emph{arXiv preprint}.

\bibitem[{Moro et~al.(2014)Moro, Raganato, and Navigli}]{moro-etal-2014-entity}
Andrea Moro, Alessandro Raganato, and Roberto Navigli. 2014.
\newblock \href {https://doi.org/10.1162/tacl_a_00179} {Entity linking meets
  word sense disambiguation: a unified approach}.
\newblock \emph{Transactions of the Association for Computational Linguistics},
  2:231--244.

\bibitem[{Moussallem et~al.(2017)Moussallem, Usbeck, Röder, and {Ngonga
  Ngomo}}]{MAG}
Diego Moussallem, Ricardo Usbeck, Michael Röder, and Axel-Cyrille {Ngonga
  Ngomo}. 2017.
\newblock Mag: A multilingual, knowledge-base agnostic and deterministic entity
  linking approach.
\newblock In \emph{K-CAP 2017: Knowledge Capture Conference}, page~8. ACM.

\bibitem[{Mulang’ et~al.(2020)Mulang’, Singh, Prabhu, Nadgeri, Hoffart, and
  Lehmann}]{Mulang__2020}
Isaiah~Onando Mulang’, Kuldeep Singh, Chaitali Prabhu, Abhishek Nadgeri,
  Johannes Hoffart, and Jens Lehmann. 2020.
\newblock \href {https://doi.org/10.1145/3340531.3412159} {Evaluating the
  impact of knowledge graph context on entity disambiguation models}.
\newblock In \emph{Proceedings of the 29th ACM International Conference on
  Information \& Knowledge Management}, CIKM ’20, page 2157–2160. ACM.

\bibitem[{Nuzzolese et~al.(2015)Nuzzolese, Gentile, Presutti, Gangemi,
  Garigliotti, and Navigli}]{Nuzzolese2015OpenKE}
Andrea~Giovanni Nuzzolese, Anna~Lisa Gentile, Valentina Presutti, Aldo Gangemi,
  Dar{\'i}o Garigliotti, and Roberto Navigli. 2015.
\newblock \href {https://api.semanticscholar.org/CorpusID:6169907} {Open
  knowledge extraction challenge}.
\newblock In \emph{SemWebEval@ESWC}.

\bibitem[{Oliveira et~al.(2021)Oliveira, Fileto, Speck, Garcia, Moussallem, and
  Lehmann}]{OLIVEIRA2021101624}
Italo~L. Oliveira, Renato Fileto, René Speck, Luís~P.F. Garcia, Diego
  Moussallem, and Jens Lehmann. 2021.
\newblock \href {https://doi.org/10.1016/j.is.2020.101624} {Towards holistic
  entity linking: Survey and directions}.
\newblock \emph{Information Systems}, 95:101624.

\bibitem[{Parravicini et~al.(2019)Parravicini, Patra, Bartolini, and
  Santambrogio}]{parravicini}
Alberto Parravicini, Rhicheek Patra, Davide~B. Bartolini, and Marco~D.
  Santambrogio. 2019.
\newblock \href {https://doi.org/10.1145/3327964.3328499} {Fast and accurate
  entity linking via graph embedding}.
\newblock In \emph{Proceedings of the 2nd Joint International Workshop on Graph
  Data Management Experiences \& Systems (GRADES) and Network Data Analytics
  (NDA)}, GRADES-NDA'19, New York, NY, USA. Association for Computing
  Machinery.

\bibitem[{Raiman and Raiman(2018)}]{tplink}
Jonathan Raiman and Olivier Raiman. 2018.
\newblock \href {https://arxiv.org/abs/1802.01021} {Deeptype: Multilingual
  entity linking by neural type system evolution}.
\newblock \emph{Preprint}, arXiv:1802.01021.

\bibitem[{R{\"o}der et~al.(2014)R{\"o}der, Usbeck, Hellmann, Gerber, and
  Both}]{roder-etal-2014-n3}
Michael R{\"o}der, Ricardo Usbeck, Sebastian Hellmann, Daniel Gerber, and
  Andreas Both. 2014.
\newblock \href
  {http://www.lrec-conf.org/proceedings/lrec2014/pdf/856_Paper.pdf}
  {N{\mbox{$^3$}} - a collection of datasets for named entity recognition and
  disambiguation in the {NLP} interchange format}.
\newblock In \emph{Proceedings of the Ninth International Conference on
  Language Resources and Evaluation ({LREC}'14)}, pages 3529--3533, Reykjavik,
  Iceland. European Language Resources Association (ELRA).

\bibitem[{R{\"{o}}der et~al.(2018)R{\"{o}}der, Usbeck, and Ngomo}]{gerbil}
Michael R{\"{o}}der, Ricardo Usbeck, and Axel{-}Cyrille~Ngonga Ngomo. 2018.
\newblock \href {https://doi.org/10.3233/SW-170286} {{GERBIL} - benchmarking
  named entity recognition and linking consistently}.
\newblock \emph{Semantic Web}, 9(5):605--625.

\bibitem[{Sevgili et~al.(2022)Sevgili, Shelmanov, Arkhipov, Panchenko, and
  Biemann}]{el_survey}
Özge Sevgili, Artem Shelmanov, Mikhail Arkhipov, Alexander Panchenko, and
  Chris Biemann. 2022.
\newblock \href {https://doi.org/10.3233/sw-222986} {Neural entity linking: A
  survey of models based on deep learning}.
\newblock \emph{Semantic Web}, 13(3):527–570.

\bibitem[{Steinmetz and Sack(2013)}]{DBLP:conf/esws/SteinmetzS13}
Nadine Steinmetz and Harald Sack. 2013.
\newblock \href {https://doi.org/10.1007/978-3-642-38288-8\_26} {Semantic
  multimedia information retrieval based on contextual descriptions}.
\newblock In \emph{The Semantic Web: Semantics and Big Data, 10th International
  Conference, {ESWC} 2013, Montpellier, France, May 26-30, 2013. Proceedings},
  volume 7882 of \emph{Lecture Notes in Computer Science}, pages 382--396.
  Springer.

\bibitem[{van Hulst et~al.(2020)van Hulst, Hasibi, Dercksen, Balog, and
  de~Vries}]{rel}
Johannes~M. van Hulst, Faegheh Hasibi, Koen Dercksen, Krisztian Balog, and
  Arjen~P. de~Vries. 2020.
\newblock \href {https://doi.org/10.1145/3397271.3401416} {{REL}: An entity
  linker standing on the shoulders of giants}.
\newblock In \emph{Proceedings of the 43rd International ACM SIGIR Conference
  on Research and Development in Information Retrieval}, SIGIR '20, page
  2197–2200, New York, NY, USA. Association for Computing Machinery.

\bibitem[{Vrande\v{c}i\'{c} and Kr\"{o}tzsch(2014)}]{wikidata}
Denny Vrande\v{c}i\'{c} and Markus Kr\"{o}tzsch. 2014.
\newblock \href {https://doi.org/10.1145/2629489} {Wikidata: a free
  collaborative knowledgebase}.
\newblock \emph{Commun. ACM}, 57(10):78–85.

\bibitem[{Wang et~al.(2023)Wang, Li, Zhu, Zhang, Perera, Hang, Ma, Wang, Wang,
  Castelli et~al.}]{wang2023benchmarking}
Sijia Wang, Alexander~Hanbo Li, Henghui Zhu, Sheng Zhang, Pramuditha Perera,
  Chung-Wei Hang, Jie Ma, William~Yang Wang, Zhiguo Wang, Vittorio Castelli,
  et~al. 2023.
\newblock Benchmarking diverse-modal entity linking with generative models.
\newblock In \emph{Findings of the Association for Computational Linguistics:
  ACL 2023}, pages 7841--7857.

\bibitem[{Wu et~al.(2020)Wu, Petroni, Josifoski, Riedel, and
  Zettlemoyer}]{blink}
Ledell Wu, Fabio Petroni, Martin Josifoski, Sebastian Riedel, and Luke
  Zettlemoyer. 2020.
\newblock \href {https://doi.org/10.18653/v1/2020.emnlp-main.519} {Scalable
  zero-shot entity linking with dense entity retrieval}.
\newblock In \emph{Proceedings of the 2020 Conference on Empirical Methods in
  Natural Language Processing (EMNLP)}, pages 6397--6407, Online. Association
  for Computational Linguistics.

\bibitem[{Xin et~al.(2024)Xin, Qi, Yao, Zhu, Zeng, Bin, Hou, and
  Li}]{xin2024llmaellargelanguagemodels}
Amy Xin, Yunjia Qi, Zijun Yao, Fangwei Zhu, Kaisheng Zeng, Xu~Bin, Lei Hou, and
  Juanzi Li. 2024.
\newblock \href {https://arxiv.org/abs/2407.04020} {{LLMAEL}: Large language
  models are good context augmenters for entity linking}.
\newblock \emph{Preprint}, arXiv:2407.04020.

\bibitem[{Zhang et~al.(2021)Zhang, Hua, and
  Stratos}]{DBLP:journals/corr/abs-2110-02369}
Wenzheng Zhang, Wenyue Hua, and Karl Stratos. 2021.
\newblock \href {https://arxiv.org/abs/2110.02369} {{EntQA}: Entity linking as
  question answering}.
\newblock \emph{CoRR}, abs/2110.02369.

\bibitem[{Zwicklbauer et~al.(2016)Zwicklbauer, Seifert, and Granitzer}]{doser}
Stefan Zwicklbauer, Christin Seifert, and Michael Granitzer. 2016.
\newblock {DoSeR} - a knowledge-base-agnostic framework for entity
  disambiguation using semantic embeddings.
\newblock In \emph{The Semantic Web. Latest Advances and New Domains}, pages
  182--198, Cham. Springer International Publishing.

\end{thebibliography}

\appendix
\begin{table*}[ht!]
\centering
\caption{Comparison for different models for the expansion. The best result is highlighted in bold.}
\begin{tabular}{@{}lcccccccc|c@{}}
\hline
\textbf{Model}          & \multicolumn{1}{l}{\textbf{AIDA}} &\multicolumn{1}{l}{\textbf{MSNBC}}&\multicolumn{1}{l}{\textbf{Der}}&\multicolumn{1}{l}{\textbf{K50}}&\multicolumn{1}{l}{\textbf{R-128}}&\multicolumn{1}{l}{\textbf{R-500}}&\multicolumn{1}{l}{\textbf{OKE 2015}}&\multicolumn{1}{l}{\textbf{OKE 2016}}&\multicolumn{1}{l}{\textbf{Avg}}  \\ \hline
LLama3 70B   & \textbf{71.6}& \textbf{69.3}& \textbf{55.7}& \textbf{70.6}& \textbf{51.7}& \textbf{56.6}& \textbf{59.4}& \textbf{58.5} & \textbf{61.7}\\
LlaMA3 8B &  69.6 &67.7 &53.2& 53.3& 48.1 &54.7& 53.5 &49.1&56.2\\
LLama 2 70B &  70.2 &68.1 &51.5&57.9& 48.2 &57.4& 57.2 &50.5&57.6\\
LLama 2 7B &  70.3 &70.0 &53.0& 51.5& 47.7 &54.7& 55.6 &54.3&57.1\\
Mistral  &  70.7& 67.5& 54.2& 48.3& 47.6 &56.1 &56.1& 56.1&57.1\\
%GrailQA s-expressions& 0.31&0.35& 0.33 &0.50 \\
\hline
\end{tabular}
\label{tab:llm_models}
\end{table*}
\section{Comparison of different LLM models}
As an appendix, we evaluated, which LLM models perform best for our entity span expansion strategy. We experimented with models with different numbers of parameters. The LLama 3 model with 70 billion parameters worked best compared to all other models. Our results show, that the expansion works better the more parameters the model has. The Mistral and the LLama 3 model with 8 billion parameters achieve similar performance.
The main reason for the difference in performance is that the larger models produce more stable outputs compared to the smaller models. In the larger models, the JSON output usually has quite similar formatting over all documents. In comparison, in the smaller models, the JSON output was slightly different from document to document, which makes it hard to extract the expansions from the output. Furthermore, the output was not always complete as there were some entities missing especially in larger sequences.

\label{sec:appendix}

\end{document}